\documentclass[a4paper,twoside]{article}

\usepackage{epsfig}
\usepackage{multicol}
\usepackage{pslatex}
\usepackage{natbib}
\let\bibhang\relax 
\usepackage{apalike}
\usepackage{amsmath,amssymb,amsfonts,amstext,amsthm}
\usepackage{algorithmic}
\usepackage{graphicx}
\usepackage{booktabs}
\usepackage{tabularx}
\usepackage[caption=false]{subfig}
\usepackage[export]{adjustbox}
\usepackage{multirow}
\usepackage[utf8]{inputenc}
\usepackage{bm}
\usepackage{xfrac}
\usepackage{numprint}
\usepackage[hyphens]{url}
\usepackage{array}
\usepackage{listings}
\usepackage{verbatimbox}
\usepackage{tikz}
\usepackage[bookmarks=false]{hyperref} 
\usepackage{cleveref} 
\usepackage[inline]{enumitem}
\usepackage{subfiles}
\usepackage{todonotes}
\usepackage{comment}
\usepackage{microtype} 
\usepackage{SCITEPRESS} 

\graphicspath{{.}{../}}

\usetikzlibrary{calc}

\newcommand{\onlyinsubfile}[1]{#1}
\newcommand{\notinsubfile}[1]{}

\newcommand{\eg}{\text{e.\,g.,\ }}

\newcolumntype{L}[1]{>{\raggedright\arraybackslash} m{#1\textwidth}}
\newcolumntype{R}[1]{>{\raggedleft\arraybackslash} m{#1\textwidth}}
\newcolumntype{C}[1]{>{\centering\arraybackslash} m{#1\textwidth}}
\newcolumntype{X}[1]{>{\raggedright\arraybackslash} m{#1\columnwidth}}
\newcolumntype{Y}[1]{>{\raggedleft\arraybackslash} m{#1\columnwidth}}
\newcolumntype{Z}[1]{>{\centering\arraybackslash} m{#1\columnwidth}}

\usepackage{xcolor}

\npthousandsep{\,}

\def\BibTeX{{\rm B\kern-.05em{\sc i\kern-.025em b}\kern-.08em
    T\kern-.1667em\lower.7ex\hbox{E}\kern-.125emX}}
    
\begin{document}

\renewcommand{\onlyinsubfile}[1]{}
\renewcommand{\notinsubfile}[1]{#1}

\title{Flexible Table Recognition and Semantic Interpretation System}

\date{}

\author{
\authorname{Marcin~Namysl\sup{1,2}\orcidAuthor{0000-0001-7066-1726},
Alexander~M.~Esser\sup{3}\orcidAuthor{0000-0002-5974-2637},
Sven~Behnke\sup{1,2}\orcidAuthor{0000-0002-5040-7525} and
Joachim~Köhler\sup{1}
}
\affiliation{\sup{1}Fraunhofer IAIS, Sankt Augustin, Germany}
\affiliation{\sup{2}Autonomous Intelligent Systems, University of Bonn, Germany}
\affiliation{\sup{3}University of Cologne, Germany}	
\email{\{first\_name\}.\{last\_name\}@iais.fraunhofer.de, aesser22@smail.uni-koeln.de}
}



\abstract
{
Table extraction is an important but still unsolved problem.
In this paper, we introduce a flexible and modular table extraction system.
We develop two rule-based algorithms that perform the complete table recognition process, including table detection and segmentation, and support the most frequent table formats.
Moreover, to incorporate the extraction of semantic information, we develop a graph-based table interpretation method.
We conduct extensive experiments on the challenging table recognition benchmarks ICDAR 2013 and ICDAR 2019, achieving results competitive with state-of-the-art approaches.
Our complete information extraction system exhibited a high $F_1$ score of 0.7380.
To support future research on information extraction from documents, we make the resources (ground-truth annotations, evaluation scripts, algorithm parameters) from our table interpretation experiment publicly available.
}

\onecolumn
\maketitle
\normalsize
\setcounter{footnote}{0}
\vfill


\section{\uppercase{Introduction}}
\label{sec:intro}

Information can hardly be presented in a more compressed way than in a table.
Humans can easily comprehend documents with tabular data.
Although automatic table extraction has been widely studied before (\S\ref{sec:related-work}), it has not yet been completely solved.
Due to the heterogeneity of document formats (\eg invoices, scientific papers, or balance sheets), this task is extremely hard.
However, as the number of digitized documents steadily increases, a solution is urgently needed to automatically extract information from tabular data.

\begin{figure*}[!htbp]
\centering
\includegraphics[width=0.9\textwidth, clip]{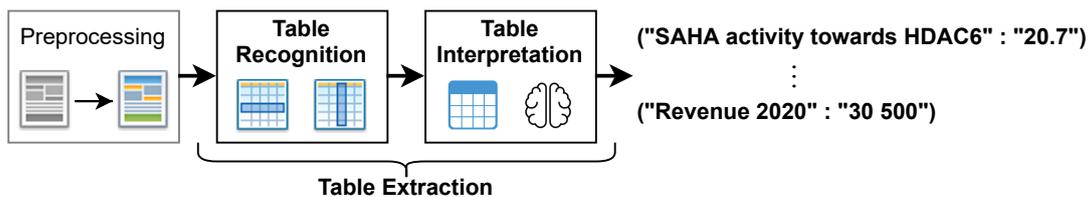}
\caption{
The diagram of our system. We focus on the table extraction task. Preprocessing involves binarization, skew angle correction, layout analysis, and OCR. Table recognition includes table detection and segmentation. Table interpretation is domain- and application-dependent.
The result is a matching between the table cells and meanings.
}
\label{fig:system-architecture}
\end{figure*}

In this paper, we introduce a holistic approach that combines table recognition and interpretation modules (\Cref{fig:system-architecture}).
Specifically, we propose two rule-based table recognition methods that perform table detection and segmentation at once. 
Our \emph{book tabs} heuristic recognizes tables that are typeset with a {\LaTeX} package \emph{booktabs}\footnote{\url{https://ctan.org/pkg/booktabs}}, which is widely used in scientific and technical publications (\S\ref{ssec:book-tabs-proposed}).
Our second algorithm handles the most popular bordered table format (\S\ref{ssec:separator-based-proposed}).
Furthermore, we complemented the basic formulation of the table recognition task by including a table interpretation module.
To this end, we implemented a rule-based interpretation method that leverages regular expressions and an
approximate (fuzzy) string matching algorithm (\S\ref{sec:interpretation-proposed}).

Our approach is \emph{flexible}, allowing both image-based and digital-born documents, and \emph{modular}, allowing us to separately adapt single processing steps. Both are crucial for a table extraction system because different processing steps need to be optimized, depending on the document type and the layout of the extremely heterogeneous input data.
For some documents, the challenge might be the table detection, for others the segmentation, or interpretation.
High interpretability of the system is essential. 
Deep learning-based end-to-end approaches, which are trained to directly extract specific values from documents, will only be suitable for specific document types contained in the training data.
Our system, however, can easily be adjusted to any document type and layout, due to its modular structure.

Our contributions can be summarized as follows:
\begin{itemize}%
\item We propose two interpretable rule-based table recognition methods developed to extract data from widely-used tabular layouts (\S\ref{sec:proposed-method}).
\item We provide a general formulation of the table interpretation task as a maximum weighted matching on a corresponding graph (\S\ref{sec:interpretation-proposed}).
\item We evaluate our table recognition algorithms on challenging data sets demonstrating the utility of our approach (\S\ref{sec:experiments-tablerec},~\S\ref{sec:experiments-tableinterpr}) and revealing issues of the evaluation protocol employed in a recent competition on table recognition.
\item We perform the evaluation of the information extraction task from tabular data and achieve a high $F_1$ score of 0.7380 proving the utility of our approach.
We make the resources from this experiment publicly available.\footnote{\url{https://github.com/mnamysl/table-interpretation}}

\end{itemize}


\section{\uppercase{Related Work}}
\label{sec:related-work}

In the following, we briefly summarize recent\footnote{Please refer to \citet{Silva2005DesignOA} for a comprehensive review of prior approaches.} approaches that perform complete table recognition.
We review both heuristic-based and learning-based methods.

\subsection{Heuristic-Based Methods}

\citet{4377094} described a system that parses the low-level data from the PDF documents and extracts the HTML representation of tables.
They locate and segment tables by analyzing the spatial features of text blocks.
Their system can detect cells that span multiple rows or columns.

\citet{Oro2009} introduced \emph{PDF-TREX}, a heuristic bottom-up approach for table recognition in single-column PDF documents. 
It uses the spatial features of page elements to align and group them into paragraphs and tables.
Similarly, it finds the rows and columns and obtains table cells from their intersections.

\citet{Nurminen2013} proposed a set of heuristics for table detection and segmentation. 
Specifically, they locate subsequent text boxes with common left, middle, or right coordinates and assign them the probability of belonging to a table object.

\citet{10.1145/2682571.2797069} proposed \emph{TEXUS}, a task-based table processing method.
They located table lines and used transitions between them and main text lines to detect table positions. 
They used alignments of text chunks inside the table region to identify columns and determined the dominant table line pattern to find rows.

\citet{10.1007/978-3-319-99972-2_20} proposed \emph{TabbyPDF}, a heuristic-based approach for table detection and structure recognition from untagged PDF documents.
Their system uses both textual and graphical features such as horizontal and vertical distances, fonts, and rulings.
Moreover, they exploit the feature of the appearance of text printing instructions and the positions of a drawing cursor.

\subsection{Learning-Based Methods}

Recently, many deep learning-based methods were proposed to solve the image-based table recognition problem.
To achieve acceptable results, these approaches need many examples for training.
Deep learning methods are often coupled with heuristics that implement the missing functionality.

\citet{Schreiber2017} proposed \emph{DeepDeSRT} that employs the \emph{Faster R-CNN model} for table detection and a \emph{semantic segmentation} approach for structure recognition.
As preprocessing, they stretch the images vertically and horizontally to facilitate the separation of rows and columns by the model.
Moreover, they apply postprocessing to fix problems with spurious detections and conjoined regions.

\citet{8893125} applied conditional \emph{Generative Adversarial Networks} for table localization and an encoder decoder-based model for table row- and column segmentation.
Segmentation was evaluated separately for rows and columns on proprietary data.

\citet{8978013} proposed \emph{TableNet}, an encoder decoder-based neural architecture for table recognition.
Their encoder is shared between the table region detection and column segmentation decoders.
Subsequently, rule-based row extraction is employed to segment individual cells.

\citet{9151030} proposed \emph{CascadeTab-Net} that uses the instance segmentation technique to detect tables and segment the cells in a single inference step.
Their model predicts the segmentation of cells only for the borderless tables and employs simple rule-based text and line detection heuristics for extracting cells from bordered tables.

In contrast to prior work, our method is flexible. It selects the required processing steps as needed and works with both image-based and digital-born PDF files.

\subsection{Table Interpretation}

Table interpretation is strongly use case-specific. There is no state-of-the-art method that fits all scenarios, but a variety of approaches from the area of natural language processing are used.

Popular methods involve string matching, calculating the Levenshtein distance~\citep{levenshtein1966bcc}, or Regular Expressions (RegEx) \citep{Kleene1951}, \eg for matching a column title or the data type of a column~\citep{Yan2018}.
Other methods, like word embeddings \citep{Mikolov2013}, entity recognition, relation extraction, or semantic parsing, semantically represent table contents.
More complex solutions are specifically trained for a certain use case, \eg  a deep learning approach to understand balance sheets.

\subsubsection{Semantic Type Detection}
The most related task to our use case is \emph{semantic type detection}. 
Semantic types describe the data by providing the correspondence between the columns and the real-world concepts, such as locations, names, or identification numbers.

A widely adopted method of detecting semantic types is to employ dictionary lookup and RegEx matching of column headers and values.
Many popular data preparation and visualization tools\footnote{Popular data analysis systems: \url{https://powerbi.microsoft.com}, \url{https://www.trifacta.com}, \url{https://datastudio.google.com}.} incorporate this technique to enhance their data analysis capabilities.

A deep learning-based approach that pairs column headers with 78 semantic types from a knowledge base was introduced recently~\citep{10.1145/3292500.3330993}. 
Moreover, \citet{zhang2020sato} combined deep learning, topic modeling, structured prediction, and the context of a column for recognition.

In favor of flexibility, we utilized the RegEx and soft string matching algorithms to detect semantic types using the content of both header and data cells.

\onlyinsubfile{
\bibliographystyle{../apalike}
\bibliography{../paper}
}


\section{\uppercase{Proposed Table Recognition Method}}
\label{sec:proposed-method}

\Cref{fig:system-architecture} illustrates the information flow between the modules of our system.
Preprocessing enables us to work with either born-digital PDF files or documents scans. 
To our best knowledge, there are few table recognition approaches that support both types of input. Most approaches require born-digital PDFs.

More specifically, we employ a layout analysis module~\citep{konya2013adaptive} to extract solid separators (ruling lines), textual, and non-textual page regions from an input document.
We then use our heuristics to extract both the location and the structure of each table.
Our methods can be easily applied to both horizontal and vertical page layouts.
In the following, we describe how they work in the case of the horizontal layout.
For the vertical layout, all steps are identical, except that we swap the horizontal and the vertical separators with each other.
Moreover, we discard all candidates that overlap any valid table region that was already detected by the previously applied heuristic.
Therefore, the order in which we apply our methods impacts the final results.
As the book tabs heuristic could generate spurious candidates from bordered tables, we first apply the separator-based method followed by the book tabs algorithm in all experiments.

Although our table segmentation algorithms would handle ruleless table layouts (cf.~\ref{ssec:book-tabs-proposed}), in this work, we are focused on information extraction from the tabular layouts that contain at least partial rulings.

%
%

\subsection{Separator-Based Table Recognition}
\label{ssec:separator-based-proposed}

{
\newcommand{\FigWidth}{0.48\textwidth}
\begin{figure}[!h]
\captionsetup[subfloat]{farskip=2pt,captionskip=0pt}
\centering
\subfloat[Input Image]{\includegraphics[width=\FigWidth]{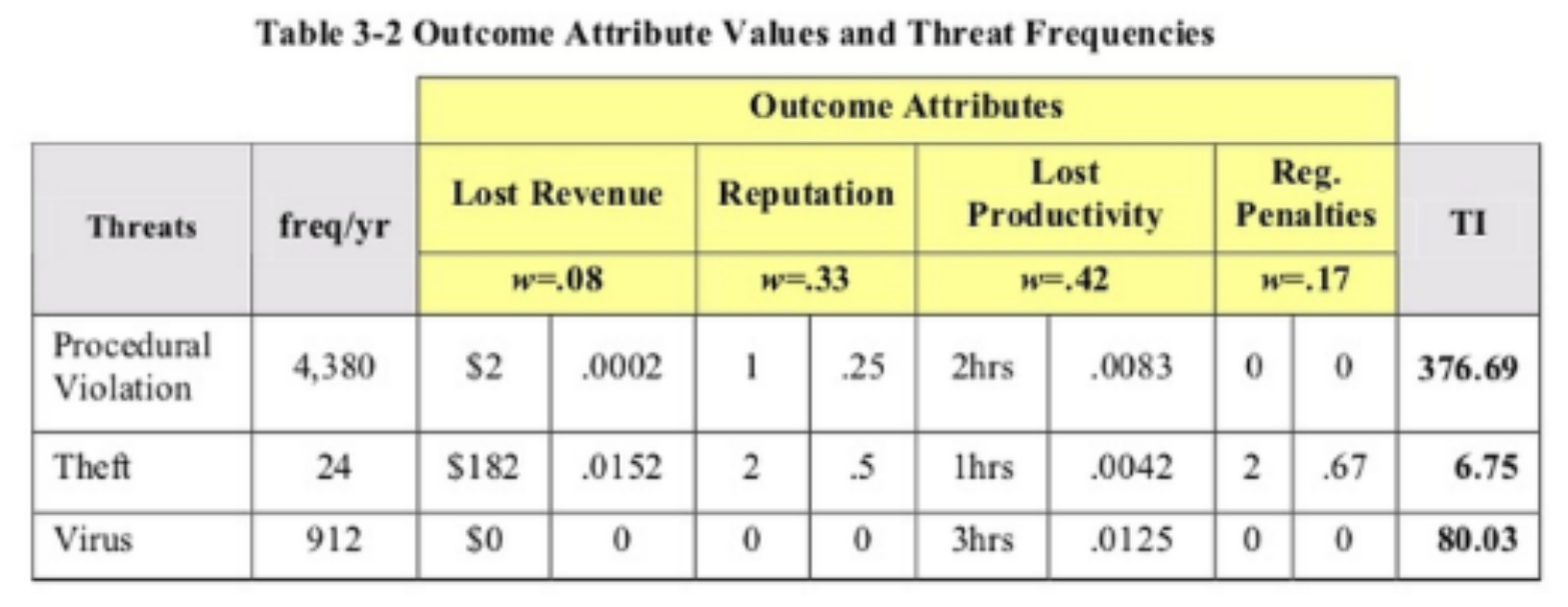}}
\hfil
\subfloat[Separator Merging\label{fig:solid-sep:sep-merging}]{\includegraphics[width=\FigWidth]{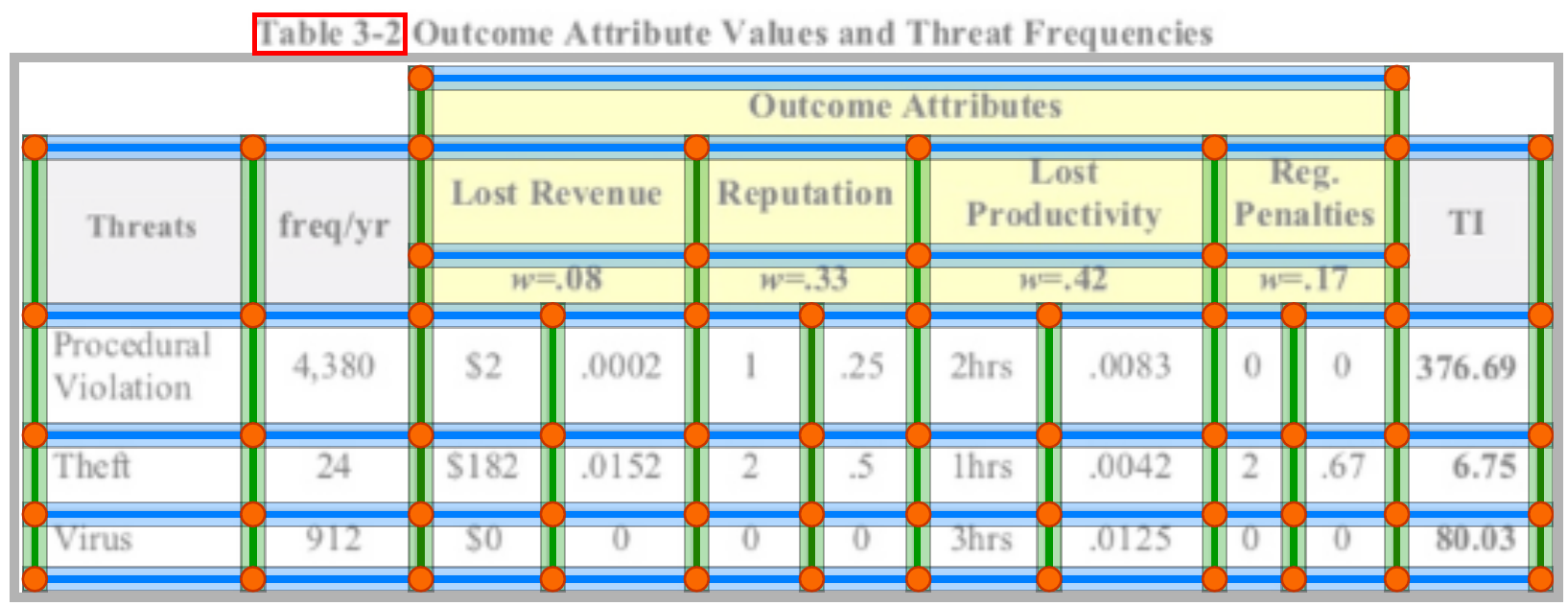}}\\
\subfloat[Cell Merging\label{fig:solid-sep:cell-merging}]{\includegraphics[width=\FigWidth]{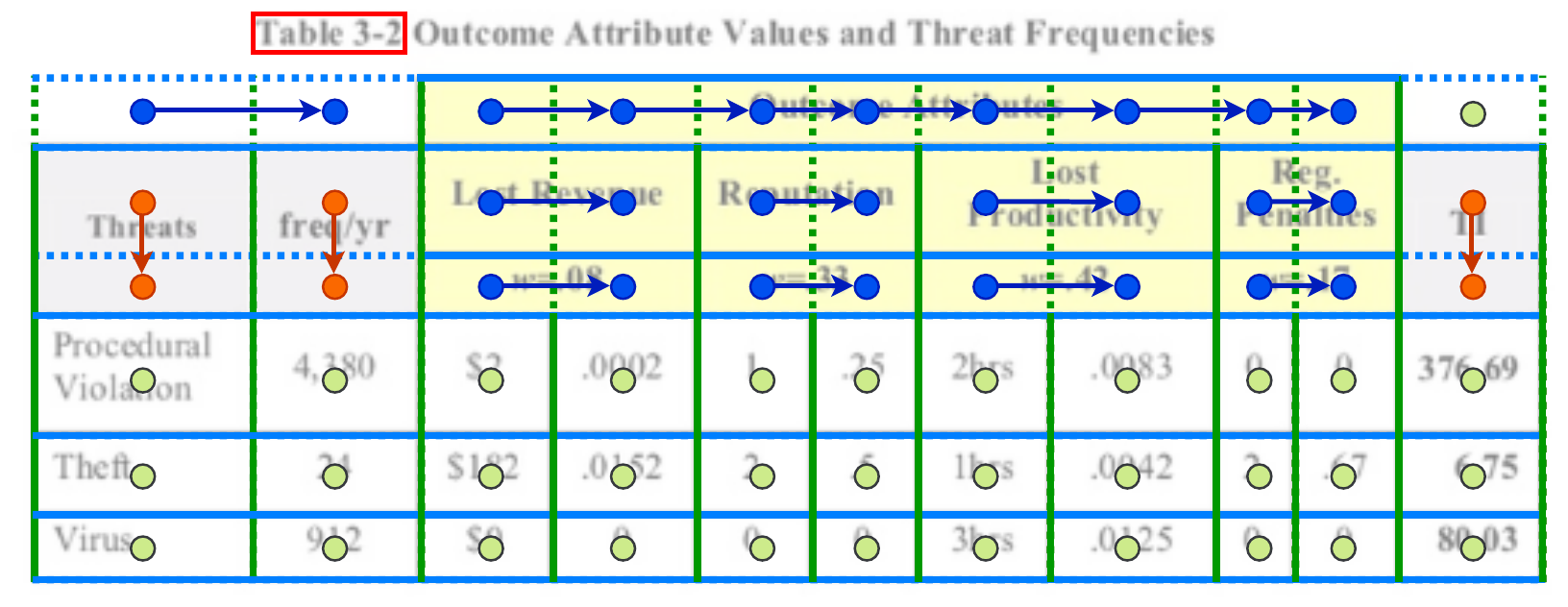}}
\hfil
\subfloat[Segmentation Result\label{fig:solid-sep:segm-res}]{\includegraphics[width=\FigWidth]{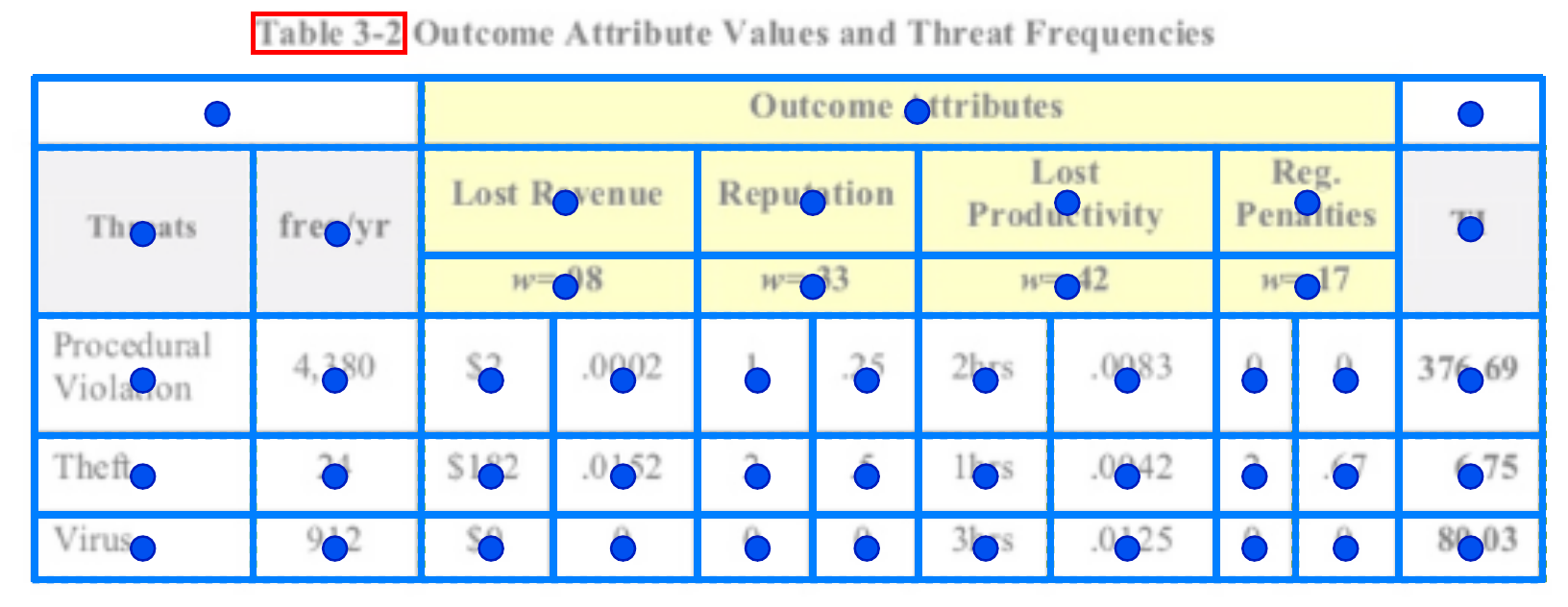}}
\par\smallskip 
\caption{
Illustration of the main processing steps of our solid separator-based heuristic:
\textbf{(a)} A table image cropped from the \emph{cTDaR\_t10047} file (ICDAR 2019).
\textbf{(b)} Separator merging stage.
Vertical and horizontal separator regions are marked green and blue, respectively. Orange circles correspond to the intersection points. The red box represents the detected table label.
\textbf{(c)} Cell merging stage.
Blue and orange circles correspond to the centers of the cells that were merged horizontally and vertically, respectively. Green circles represent the center of fully bordered cells. Arrows mark the merging direction.
\textbf{(d)} Segmentation result.
Blue circles represent the center of the recognized cells.
}
\label{fig:example-solid-sep}
\end{figure}
}

Our separator-based heuristic (\Cref{fig:example-solid-sep}) starts by sorting the horizontal and the vertical separators by the top and the left position, respectively.

\subsubsection{Separator Merging}
First, all separator boxes are expanded by 5 pixels on each side to increase the chance of intersection with the neighboring ruling lines that have a different orientation (vertical or horizontal).
Then, we iteratively merge all separator boxes if they intersect with each other (\Cref{fig:solid-sep:sep-merging}).
We repeat this process until no further intersection can be found.
\subsubsection{Table Labels Assignment}
To improve precision, we search for the presence of predefined keywords (\eg "table", "Tab.") in the close neighborhood of separator boxes and mark the table as \emph{labeled} or \emph{unlabeled}.
If the labels are required by the current configuration, we remove all unlabeled tables from the set of already found candidates.

\subsubsection{Table Grid Estimation}
Subsequently, we derive a rough grid structure of each table candidate.
Each pair of subsequent vertical or horizontal separators forms a table column or table row region, respectively.
We calculate the cell regions based on intersections between the column and the row boxes.
This procedure returns a list of roughly segmented table grids.
\subsubsection{Table Grid Refinement}
Some cells in the grid need to be refined by merging them with the neighboring cells to recover the multi-row and multi-column cells. 
Our approach is inspired by the \emph{union-finding algorithm}~\citep{PhysRevB.14.3438}.
We perform a raster scan through the rough grid of cells.
For each cell, scanning in the left-to-right direction, we check whether the area near the right border of the cell's box overlaps any vertical separator assigned to the current table candidate.
If this is not the case, we merge the current cell with its right neighbor and proceed to the next cell.
This procedure is then repeated in the top-to-down direction.
We illustrate this process in \Cref{fig:solid-sep:cell-merging}.
Note that the column spans of the cells that need to be merged must be equal.

%
%

\subsection{Book Tabs Table Recognition}
\label{ssec:book-tabs-proposed}

{
\newcommand{\FigWidth}{0.48\textwidth}
\begin{figure}[!htbp]
\captionsetup[subfloat]{farskip=1pt,captionskip=1pt}
\centering
\subfloat[Input Image\label{fig:book-tabs:input1}]{\includegraphics[width=\FigWidth,clip,trim=0 0 0 0]{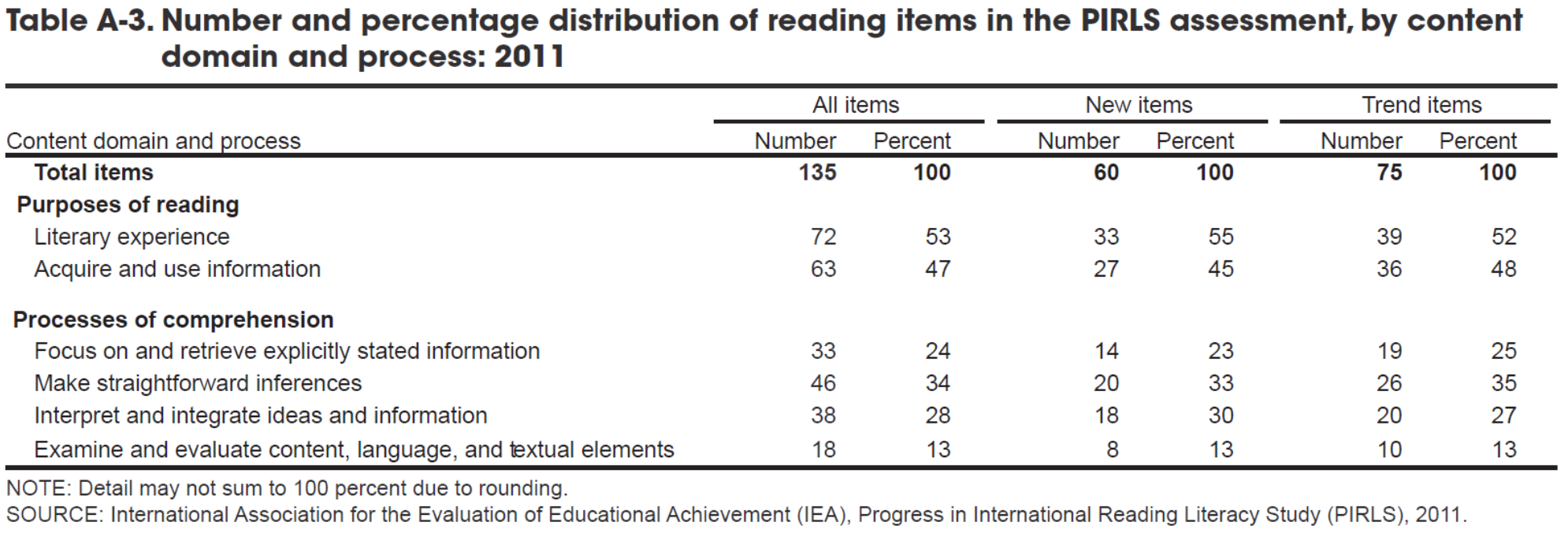}}
\hfil
\subfloat[Segmentation Grid\label{fig:book-tabs:segm-grid}]{\includegraphics[width=\FigWidth]{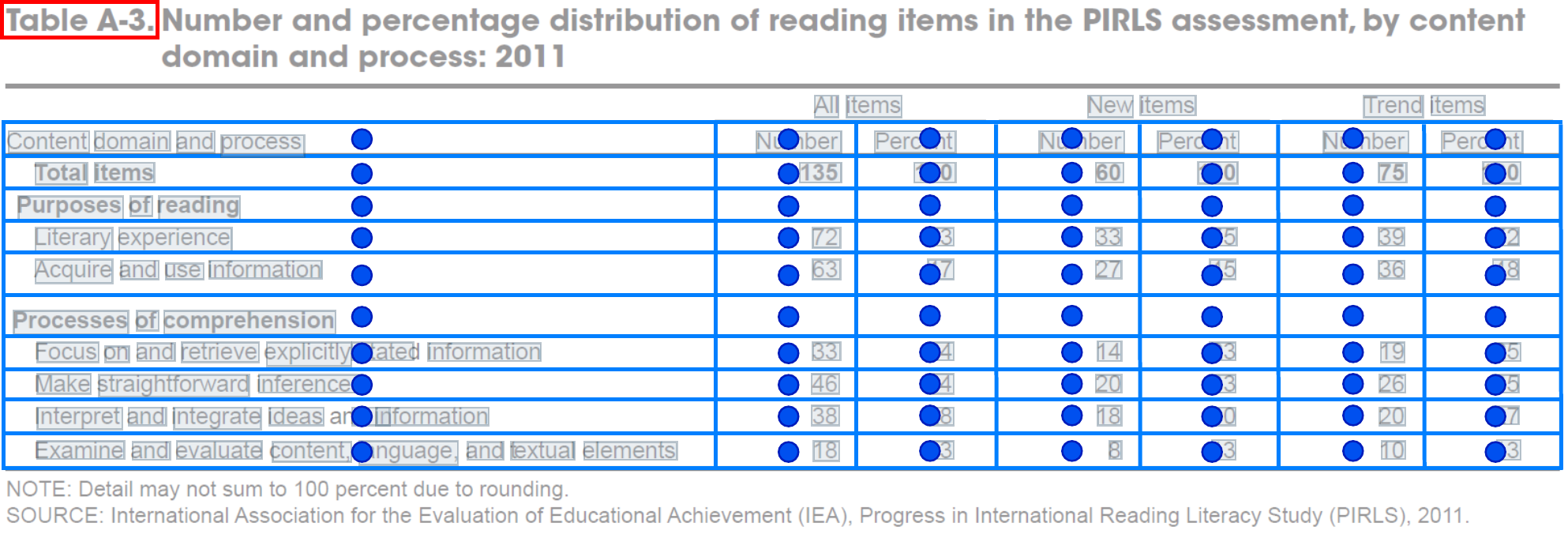}}
\hfil
\subfloat[Row Segmentation\label{fig:book-tabs:row-segm}]{\includegraphics[width=\FigWidth]{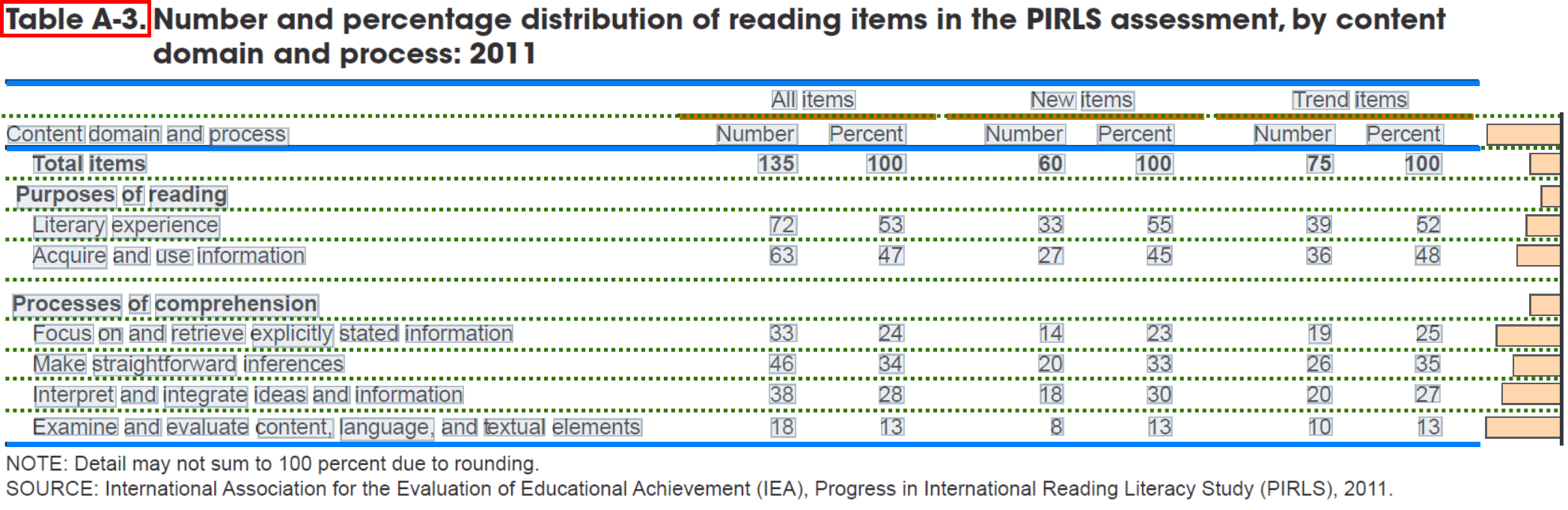}}
\hfil
\subfloat[Column Segmentation\label{fig:book-tabs:col-segm}]{\includegraphics[width=\FigWidth]{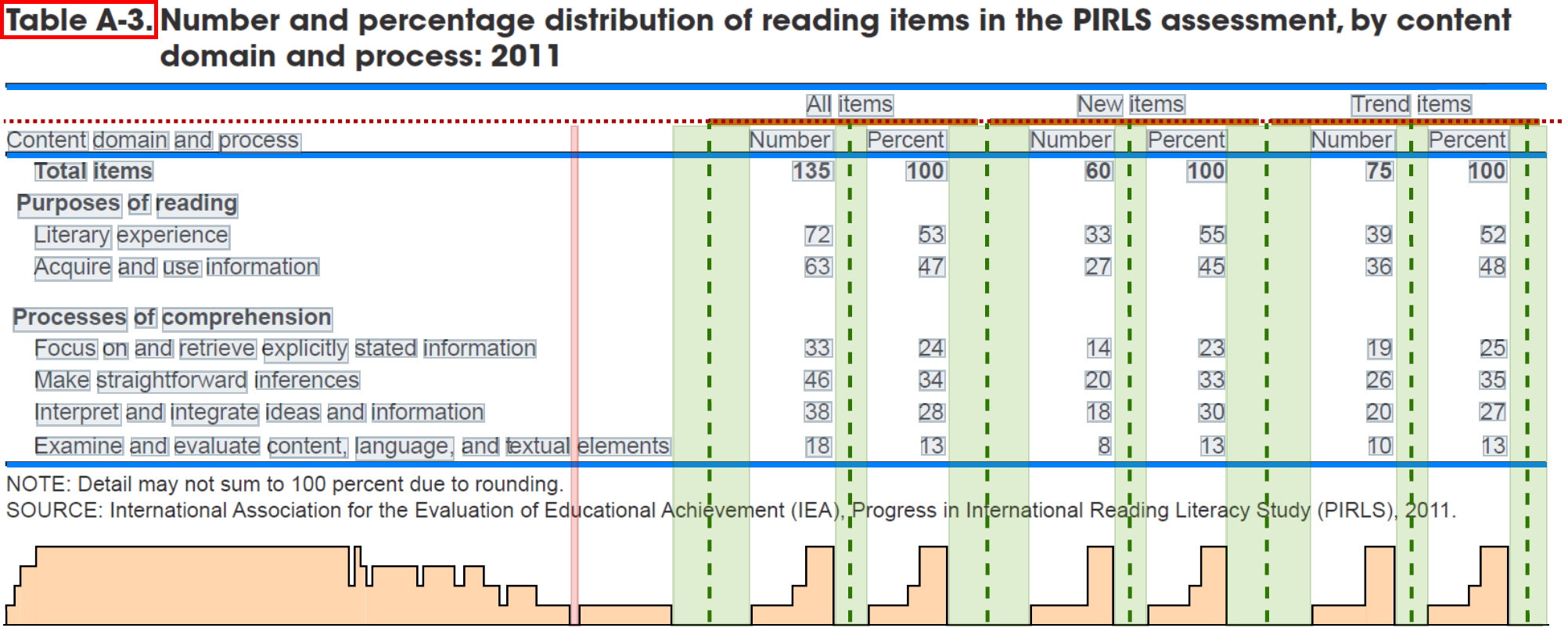}}
\par\smallskip 
\caption{
Illustration of the table body and lowest-level header segmentation of our book tabs-based heuristic.
\textbf{(a)} A table image cropped from the \emph{us-021} file (ICDAR 2013).
\textbf{(b)} The resulting segmentation grid. 
Blue lines and circles represent the borders and the centers of the cells, respectively.
The boxes with gray backgrounds outline the words within the table area.
\textbf{(c)} Row segmentation. 
Blue and orange lines represent the top/middle/bottom and the \emph{cmidrule} rule lines, respectively. 
Orange bars to the right correspond to the horizontal profile (running sum of pixels in the text regions in each row).
Green dotted lines correspond to the row borders.
\textbf{(d)} Column segmentation.
The dotted red line is a border of the lowest-level header.
Orange bars at the bottom correspond to the vertical profile (running sum of pixels in the word regions in each column).
We clip the values in the profile for better visualization.
The column gaps that are wider/narrower than $\mathcal{D}_\text{column}$ are highlighted in green/red, respectively. 
Green vertical dotted lines represent the detected column borders.
}
\label{fig:example-book-tabs-body}
\end{figure}
}

The book tabs format consists of three main components: a top, middle, and bottom rule (cf. examples in \Cref{fig:book-tabs:input1} and \Cref{fig:book-tabs:input2}).
The middle rule separates the table header and the table body region.
Optionally, a multi-level header structure can be represented using shorter \emph{cmidrules}.
These inner rules span multiple columns aggregated under the same higher-level header.
Our book tabs heuristic (\Cref{fig:example-book-tabs-body} and \Cref{fig:example-book-tabs-header}) uses horizontal separators for documents with standard orientation. 

\subsubsection{Table Region Detection}
First, the separators are sorted by the top position.
We search for triples of consecutive separators with similar coordinates of their left and right sides.
Moreover, we perform a label assignment step as in §\ref{ssec:separator-based-proposed}.

\subsubsection{Table Rows Detection}
In the previous step, we also collect all \emph{inner separators} (cmidrules) that are located between the top and the middle rule.
We group all inner separators by their \emph{y} position to isolate different levels of the header's hierarchy and to separate header rows.
The row borders in the body region of the table are determined using the horizontal profile calculated by projecting all words within the body region of a table candidate (\Cref{fig:book-tabs:row-segm}). 
The row borders are then estimated in the middle of the gaps in the resulting profile.

\subsubsection{Table Columns Detection}
We determine the median unit distance $\mathcal{D}_\text{page}$ (the distance divided by the word height) between two words within a page. 
Assuming that the table font is constant, we also calculate the mean word height within the \mbox{table} $\mathcal{H}_\text{table}$.
For each table candidate, we determine the threshold used to locate gaps between two consecutive table columns using $\mathcal{D}_\text{column} = \mathcal{D}_\text{page} \mathcal{H}_\text{table} \gamma$, where $\gamma$ is a hyperparameter.
We project all page regions within the body region and the lowest-level header row (\Cref{fig:book-tabs:col-segm}).
The higher-level headers are excluded, as they contain multi-column cells.
We analyze the projection to find all intervals with a length above $\mathcal{D}_\text{column}$.
The center positions of these intervals form the column borders.
All gaps with length below $\mathcal{D}_\text{column}$ correspond to vertically aligned words that form spurious columns.

\subsubsection{Table Grid Estimation and Refinement}
We compute the grid of the cell boxes from the intersections between the row and the column borders (\Cref{fig:book-tabs:segm-grid}).
To recognize the complex header hierarchy, we merge all roughly detected cells that intersect the same inner separator segment (\Cref{fig:book-tabs:header-merge}).

{
\newcommand{\FigWidth}{0.78\textwidth}
\begin{figure*}[!htbp]
\captionsetup[subfloat]{farskip=1pt,captionskip=1pt}
\centering
\subfloat[Input Image\label{fig:book-tabs:input2}]{\includegraphics[width=\FigWidth,clip,trim=0 2cm 0 0]{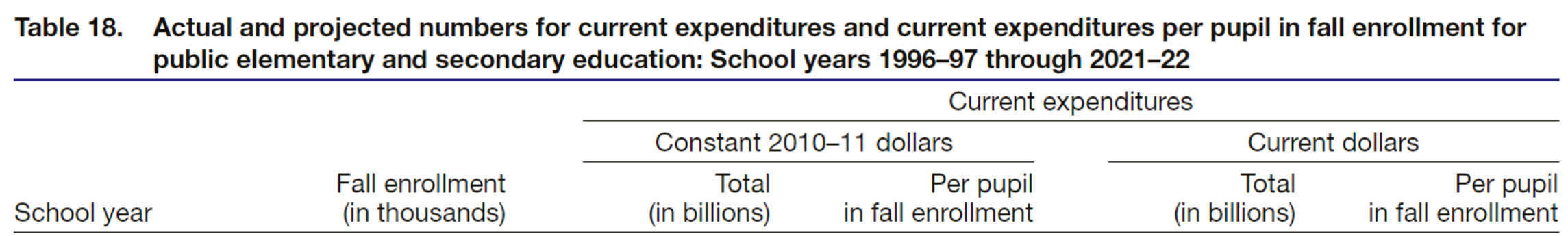}}\\
\subfloat[Header Cell Merging\label{fig:book-tabs:header-merge}]{\includegraphics[width=\FigWidth,clip,trim=0 2cm 0 0]{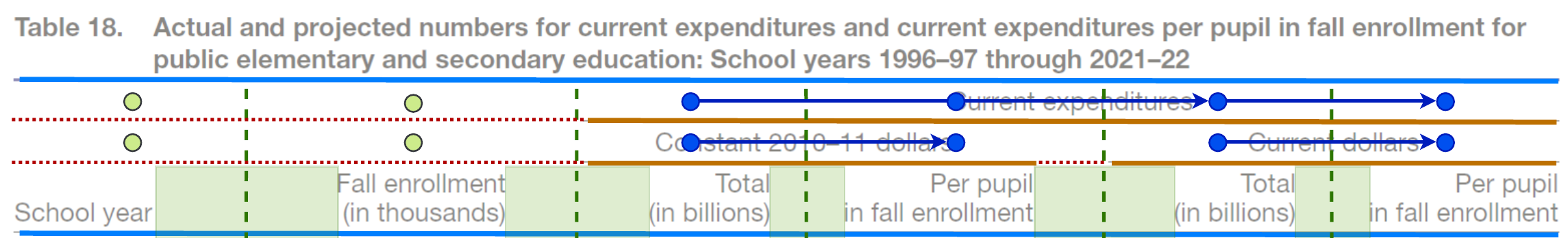}}\\
\subfloat[Final Header Segmentation\label{fig:book-tabs:header-final}]{\includegraphics[width=\FigWidth,clip,trim=0 2cm 0 0]{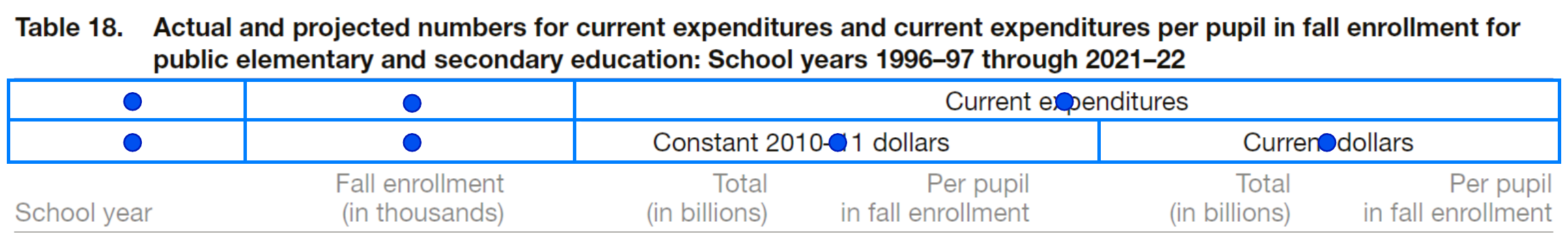}}
\par\smallskip 
\caption{
Illustration of the higher-level header segmentation of our book tabs-based heuristic.
\textbf{(a)} The top part of a table from the \emph{us-018} file (ICDAR 2013).
\textbf{(b)} Header cell merging. 
Orange lines represent the \emph{cmidrule} lines.
Green areas and lines correspond to column white spaces and borders, respectively.
Blue circles are the centers of the cells that intersect the same cmidrule line and thus need to be merged.
The cells marked with green circles remain unchanged.
\textbf{(c)} Header segmentation.
Blue lines and circles correspond to the borders and the centers of the cells in the final grid, respectively.
}
\label{fig:example-book-tabs-header}
\end{figure*}
}
%


\section{\uppercase{Proposed Table Interpretation Method}}
\label{sec:interpretation-proposed}

\begin{figure*}[!htbp]
\captionsetup[subfloat]{farskip=1pt,captionskip=1pt}
\centering
\subfloat[Input Table]{\fbox{
\begin{adjustbox}{valign=c}
\begin{tikzpicture}
\node[anchor=north west,inner sep=0] at (0,0) {\includegraphics[width=0.5\textwidth, clip, trim=2.25cm 15.25cm 10.8cm 9.85cm, page=5]{{../figures/external/13_2019.01055}.pdf}};
\draw[blue,thick,rounded corners] (3.9,-1.3) rectangle (5.1,-3.45);
\draw[blue,thick,rounded corners] (-0.1,-1.10) rectangle (1.4,-3.45);
\end{tikzpicture}
\end{adjustbox}
}}
\hfil
\subfloat[Configuration File\label{fig:interpretation_config}]{
\begin{adjustbox}{valign=c}
\verbfilebox[\tiny\hspace{0ex}]{../figures/external/config.txt}\fbox{\theverbbox}
\end{adjustbox}
}
\hfil
\subfloat[Interpretation Graph\label{fig:interpretation_graph}]{
\includegraphics[width=0.44\textwidth, valign=c]{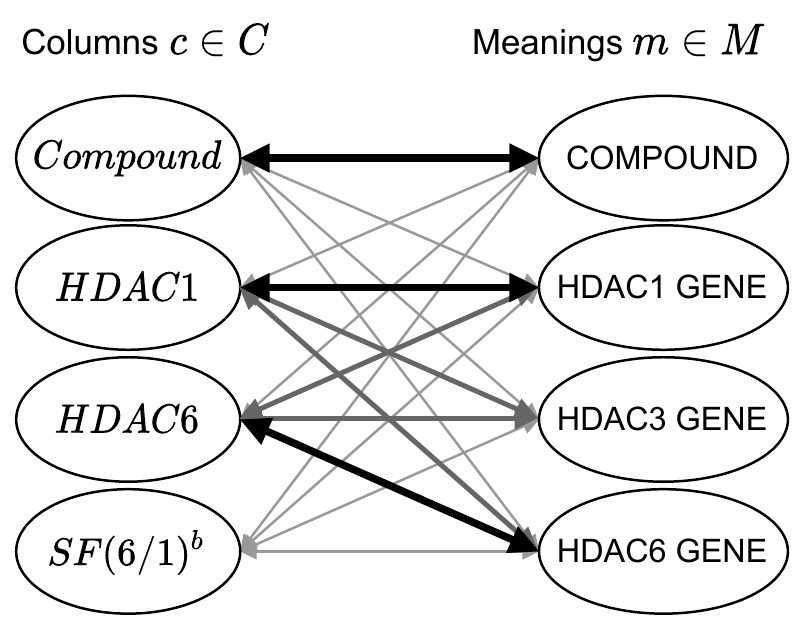}
}
\hfil
\subfloat[Extracted Tuples\label{fig:interpretation_result}]{
\begin{adjustbox}{valign=c}
\verbfilebox[\scriptsize\hspace{0ex}]{../figures/external/tuples.txt}\fbox{\theverbbox}
\end{adjustbox}
}
\par\smallskip 
\caption{
Illustration of our table interpretation method:
\textbf{(a)} A table extracted from \citet{Miao2019}.
The columns corresponding to the defined meanings are marked with blue boxes.
\textbf{(b)}~A~JSON file defining the meanings \uppercase{compound} and \uppercase{HDAC6 gene}, and rules for matching columns to these meanings.
\textbf{(c)}~Table interpretation graph: Columns $c\in C$ are mapped to the meanings $m\in M$.
For each mapping, an affinity value is calculated, symbolized by the thickness of the arrows.
\textbf{(d)}~The extracted tuples that represent the inhibitory activity of each compound towards the HDAC6 gene.}
\label{fig:interpretation-method}
\end{figure*}

Our algorithm takes a recognized table as input and assigns meanings $m\in M$ to the columns $c \in C$ (\Cref{fig:interpretation_graph}). 
For each $m$, we define a customized set of affinity rules describing a column that is likely to be matched with $m$ (\Cref{fig:interpretation_config}): 
\begin{enumerate}[label=(\arabic*)]
	\item \emph{Title Keyword Score}: Fuzzy matching between the title of a column and the predefined keywords.
	\item \emph{Title RegEx Score}: Exact matching of the title of a column with a customized regular expression.
        \item \emph{Data Type Score}: Exact matching of the content of the cells in a column with regular expressions for predefined types (e.g., integer, date, etc.).
      	\item \emph{Content RegEx Score}: Exact matching of the content of the cells in a column with a customized regular expression.
\end{enumerate}
Fuzzy matching corresponds to the Levenshtein~\citep{levenshtein1966bcc} distance between two strings divided by the length of the longer string.
The RegEx scores return $1.0$ if the matching succeeded and $0.0$ otherwise.
Moreover, the content and data type scores are averaged over the scores for the cells in the corresponding column. 
The final affinity score $S$ for a column $c$ with a meaning $m$ is computed as presented in \Cref{eq:affinity-score}:

\begin{equation}
\centering
\label{eq:affinity-score}
S(c, m) = \frac{w_\text{c} \max(S_\text{c}^\text{Rx}, S_\text{c}^\text{DT}) + w_\text{t} \max(S_\text{t}^\text{Rx}, S_\text{t}^\text{KW})}{w_\text{c} + w_\text{t}},
\end{equation}
where $w_\text{t}$ and $w_\text{c}$ are the weights of the title and the content property groups, respectively.
$S_\text{c}^\text{Rx}$ and $S_\text{c}^\text{DT}$ are the affinity scores of the content RegEx and the data type, respectively.
$S_\text{t}^\text{Rx}$ and $S_\text{t}^\text{KW}$ correspond to the scores of the title RegEx and the fuzzy matching with the keywords, respectively.
The sum of weights must be a positive number.
Moreover, if a particular rule is not defined for a meaning, the corresponding score is set to zero. 

To perform the matching between the meanings and the columns in a table, we create a weighted bipartite graph with two sets of vertices: the meanings on one side and the columns on the other side (\Cref{fig:interpretation_graph}).
We first link all columns with all meanings with an edge weighted by the affinity scores that specify how likely a column matches with a certain meaning.
We prune the connections that do not reach a predefined required minimum affinity value.
Subsequently, we find the best assignment using \emph{maximum weighted matching}~\citep{Edmonds1965b} on the bipartite graph.
Finally, we extract the tuples $x_{i,j}$, where $i$ is an index of a row in the body part of the table, and $j$ is the index of a matched meaning (cf.~\Cref{fig:interpretation_result}). 

\onlyinsubfile{
\bibliographystyle{../splncs04}
\bibliography{../paper}
}


\section{\uppercase{Table Recognition Experiments}}
\label{sec:experiments-tablerec}

\subsection{Data Sets}

The \emph{ICDAR 2013 Table Competition} data set~\citep{Goebel2013} contains born-digital business and government PDF documents with 156 tables. Ground-truth annotations for both table detection and segmentation tasks are available. As many tables in this data set are compatible with our heuristics, we refer to the experiment on this data set as the \emph{in-domain evaluation}.

The \emph{ICDAR 2019 Table Detection and Recognition} data set (cTDaR;~\citet{Gao2019}) is a collection of modern and archival document images.
We used only the former part, as the latter consists of handwritten documents and the analysis of hand-drawn tables is outside the scope of this work.
We evaluated the complete recognition process (track B2), as it is the most challenging task.
As this data set contains various tabular layouts, we regard this experiment as an \emph{out-of-domain evaluation}.

\subsection{Hyperparameters}
\label{ssec:hyperparams}
For ICDAR 2013, the table labels were required by the book tabs heuristic.
For ICDAR 2019, we fed images to the layout analysis component, and turn off the requirement of table label presence for both heuristics.
We set $\gamma=2.0$ in all experiments.
We tuned the above hyperparameters on two held-out sets: The practice data released in the ICDAR 2013 competition (no intersection with the evaluation test set) that consists of $58$ PDF documents and $16$ images randomly sampled from the \emph{track A} data set (table detection, not evaluated in this paper) of the ICDAR 2019 competition. 

%
%
\subsection{ICDAR 2013 Results (In-Domain Evaluation)}
\label{ssec:results-icdar2013}

{\renewcommand{\arraystretch}{1.0}\setlength{\tabcolsep}{5.0pt}
\newcommand{\ColWidthA}{0.3}
\newcommand{\ColWidthN}{0.14}
\begin{table*}[!ht]
\caption{
ICDAR 2013 evaluation. We report the precision, recall, and F$_{1}$ score (per-document averages) for the complete recognition process.
}
\centering
\begin{tabular}{@{}L{\ColWidthA}*{4}{C{\ColWidthN}}@{}}
\toprule
\textbf{Method} & \textbf{Precision} & \textbf{Recall} & $\mathbf{F_{1}}$\\
\midrule
FineReader~\citep{Goebel2013} & 0.8710 & \textbf{0.8835} & \textbf{0.8772}\\
OmniPage~\citep{Goebel2013} & 0.8460 & 0.8380 & 0.8420\\
Nurminen~\citep{Goebel2013} & 0.8693 & 0.8078 & 0.8374\\
Ours & \textbf{0.9179} & 0.7616 & 0.8325\\
TabbyPDF~\citep{10.1007/978-3-319-99972-2_20} & 0.8339 & 0.8298 & 0.8318\\
TEXUS~\citep{10.1145/2682571.2797069} & 0.8071 & 0.7823 & 0.7945\\
\bottomrule
\end{tabular}
\label{tab:icdar2013-results}
\end{table*}}

We first validated our approach on a popular table recognition benchmark from the ICDAR 2013 Table Competition\footnote{\url{http://www.tamirhassan.com/html/competition.html}}.
We matched the ground-truth and the recognized tables at the Intersection over Union (IoU) threshold of 0.5 ($IoU_\text{min}$).
IoU is defined as the ratio between the area of the overlap and the union of two bounding boxes.
If two tables were matched at $IoU\ge IoU_\text{min}$, their lists of \emph{adjacency relations}, i.e., relations between the neighboring cells in a table, were compared by using precision and recall measures (cf. \citet{Goebel2013}). 
All ground-truth tables that did not match with any recognized table at $IoU\ge IoU_\text{min}$ were classified as not detected and their adjacency relations were counted as false-negative relations. 
Consequently, all adjacency relations from the recognized tables that did not match with any ground-truth table at $IoU\ge IoU_\text{min}$ were counted as false-positive relations.
We included all false-positive and false-negative relations in the reported complete recognition scores (precision, recall, and F$_1$).

In \Cref{tab:icdar2013-results}, we present the results of our method in the complete recognition task.
For comparison, we present the best previously published results on this data set\footnote{We only included the prior work that reported the results of the complete recognition process. 
Moreover, we excluded methods that used a subset of the data for evaluation.}.
We outperform the other methods in terms of precision.
Moreover, we achieve the F$_{1}$ score on par with the best academic methods. 

\subsection{ICDAR 2019 Results (Out-of-Domain Evaluation)}
\label{ssec:results-icdar2019}

{\renewcommand{\arraystretch}{1.0}\setlength{\tabcolsep}{0.0pt}
\newcommand{\ColWidthA}{0.35}
\newcommand{\ColWidthN}{0.12}
\newcommand{\ColWidthB}{0.12}
\begin{table*}[htbp]
\caption{
ICDAR 2019 evaluation. We report the precision, recall, and F$_{1}$ score for the complete table recognition process for track B2 (modern documents) for two reference values of the IoU threshold between the ground-truth and the recognized cells.
For comparison, we include the best results reported in previous works.
\emph{WAvg.F$_{1}$} denotes the average F$_{1}$ score weighted by the IoU threshold for IoU $\in \{0.6, 0.7, 0.8, 0.9\}$.
}
\centering
\begin{tabular}{@{}L{\ColWidthA}*{4}{C{\ColWidthN}}C{\ColWidthB}@{}}
\toprule
\multirow{2}{\ColWidthA\textwidth}[-2pt]{\textbf{Method}} & \multicolumn{2}{c}{\textbf{IoU = 0.6}} & \multicolumn{2}{c}{\textbf{IoU = 0.7}} & \multirow{2}{\ColWidthB\textwidth}[-2pt]{\textbf{WAvg. F$_{1}$}} \\
\cmidrule(lr){2-3}\cmidrule(lr){4-5}
 & \textbf{Precision} & \textbf{Recall} & \textbf{Precision} & \textbf{Recall} & \\
\midrule
CascadeTabNet~\citep{9151030} & - & - & - & - & \textbf{0.23} \\
NLPR-PAL~\citep{Gao2019} & 0.32 & \textbf{0.42} & 0.27 & \textbf{0.35} & 0.20\\
Ours & \textbf{0.45} & 0.16 & \textbf{0.41} & 0.14 & 0.16\\
HCL IDORAN~\citep{Gao2019} & 1E-3 & 1E-3 & 1E-3 & 7E-4 & 3E-4\\
\bottomrule
\end{tabular}
\label{tab:icdar2019-results}
\end{table*}}

We tested our approach on the cTDaR data set of document scans with various layouts, employing the official tools and metrics for evaluation\footnote{\url{https://github.com/cndplab-founder/ICDAR2019_cTDaR}}. 

\Cref{tab:icdar2019-results} presents our results in comparison with the best-reported scores. 
Note that only two participant methods were registered for the structure recognition track in this competition, which emphasizes the difficulty of this task.
Our approach was outperformed by the deep learning-based methods that reported the state-of-the-art results in terms of the weighted average F$_{1}$ score (\emph{WAvg.F$_{1}$}).

\begin{figure*}[htb]
\centering
\subfloat{\includegraphics[width=0.4\textwidth, valign=t]{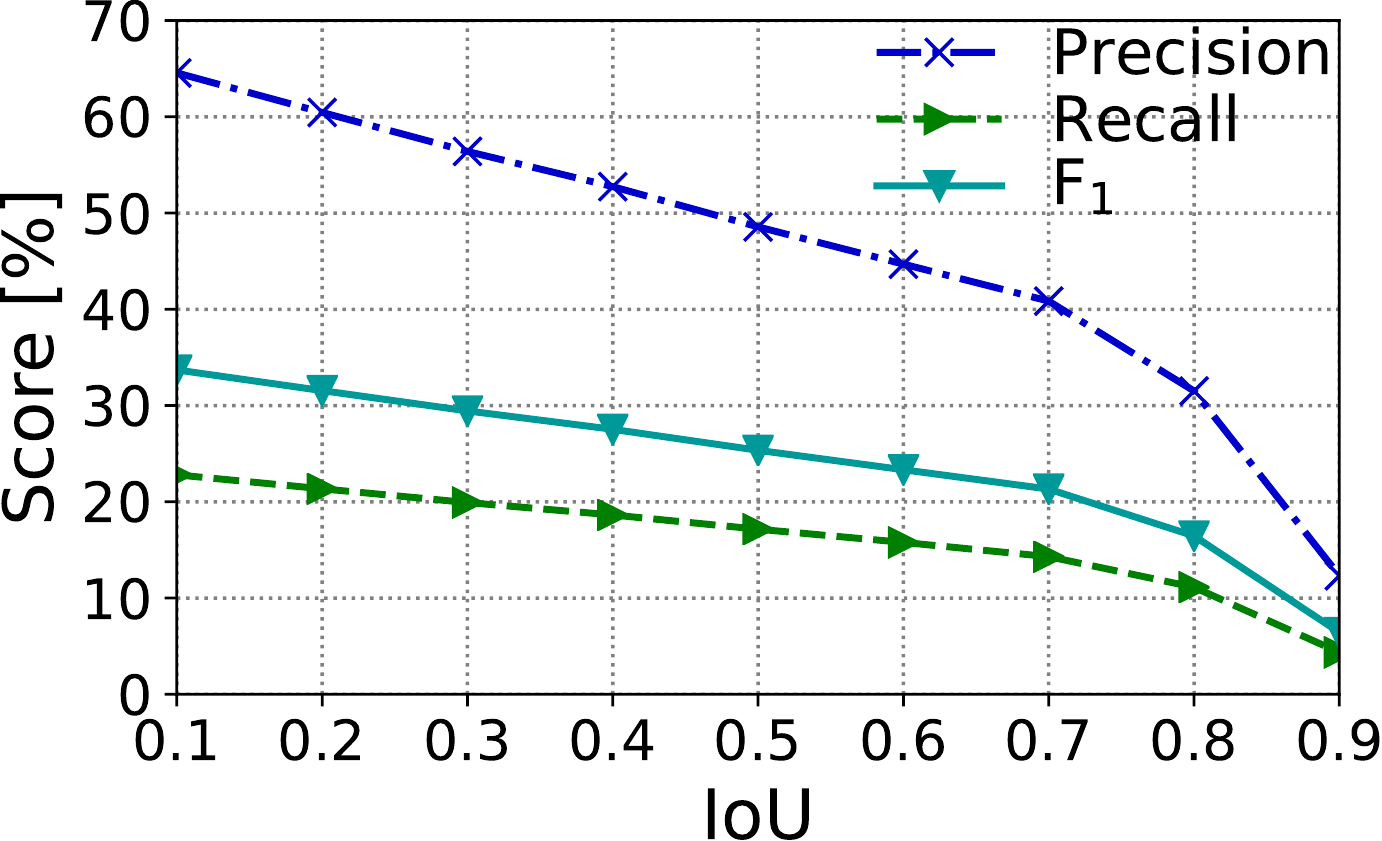} 
}\hfil
\subfloat{\includegraphics[width=0.57\textwidth, valign=t]{{../figures/icdar2019_eval/example}.png}} 
\par\smallskip 
\caption[LOF]{Extended ICDAR 2019 data set results for our method. 
\textbf{Left}: IoU-F$_{1}$, IoU-Precision, and IoU-Recall curves for the complete recognition process.
\textbf{Right}: An excerpt from the \emph{cTDaR10001} document. 
Blue boxes correspond to the ground-truth regions. 
Green and red boxes are the regions recognized above and below the IoU$=0.9$ threshold, respectively.
Above the boxes, we marked the text recognized by the Tesseract 4 OCR engine\footnotemark
depicted in the ground-truth and  recognized regions (green text indicates identical results).
The results for regions produced by our method are better, although several matches were rejected at IoU$=0.9$, which lowered our \emph{WAvg.F$_{1}$} score.
}
\label{fig:icdar2019}
\end{figure*}

Nevertheless, we argue that this score is not adequate to compare table recognition systems.
In \Cref{fig:icdar2019}, we report the detailed results of our method with different IoU thresholds for the cell matching procedure.
The \emph{WAvg.F$_{1}$} measure proposed in this competition is biased towards high overlap-ratios between cells and strongly penalizes lower IoU scores, although such scores are not necessarily needed to reliably recognize the content of the cells, which is the ultimate goal of information extraction from tabular data.
Note that our method achieved results better than the best \emph{WAvg.F$_{1}$} score for all IoU thresholds less than or equal to 0.7.
Moreover, our approach exhibits high precision, outperforming the state-of-the-art results.
The lower recall scores result from the large variety of table layouts present in this data set.

\subsection{Discussion and Limitations}
\label{ssec:limitations}

Besides many advantages, we also noticed some limitations in our approach.
First, it is prone to the errors propagated from the upstream components of our system that may cause missing or spurious separators.
Moreover, heuristic-based methods generally exhibit lower recall, as the hand-crafted rules need to be designed for each supported layout.
Heuristic methods are, however, more interpretable and can be extended to other scenarios.
Furthermore, our system is not fully parameter-free.
Due to the great heterogeneity of documents, the parameters have to be adapted to different layouts.
However, in contrast to the deep learning methods, this adaptation requires comparatively less training data.


\section{\uppercase{Table Interpretation Experiment}}
\label{sec:experiments-tableinterpr}

\footnotetext{\url{https://github.com/tesseract-ocr/tesseract}}

\subsection{Evaluation Data Set}

For table interpretation, we were not able to find any common, publicly available benchmark, neither for general data nor for our use case.
Therefore, we annotated 13 documents with tables from our biomedical data collection.
We selected documents containing tables presenting the inhibitory activity of different compounds toward the \emph{HDAC}\footnote{\emph{Histone deacetylase}: \url{https://en.wikipedia.org/wiki/Histone_deacetylase}. 
Specifically, we focused on the HDAC1 and HDAC6 target genes. 
} gene.

The ground-truth data for a table consists of a list of tuples, each representing an intersection of a data row and the columns that correspond to the defined meanings.
The annotations are stored in JSON files (cf. \Cref{fig:interpretation_result}) with the following name pattern:
\begin{verbatim}
<FILE_ID>_<PAGE_NR>_<TABLE_IDX>.json
\end{verbatim}
where \verb!<FILE_ID>! is the file identifier, \verb!<PAGE_NR>! is the page number in the corresponding PDF file, and \verb!<TABLE_IDX>! is the index of a table on a page.

We manually annotated 113 tuples from 17 tables and used them as ground-truth data in our experiment. 
We present an example of a ground-truth file in \Cref{fig:interpr-gt} in the appendix.
Moreover, we selected a separate development set of four documents for fine-tuning.

\subsection{Evaluation Setup}

\begin{figure*}[!htbp]
\centering
\includegraphics[width=0.96\textwidth]{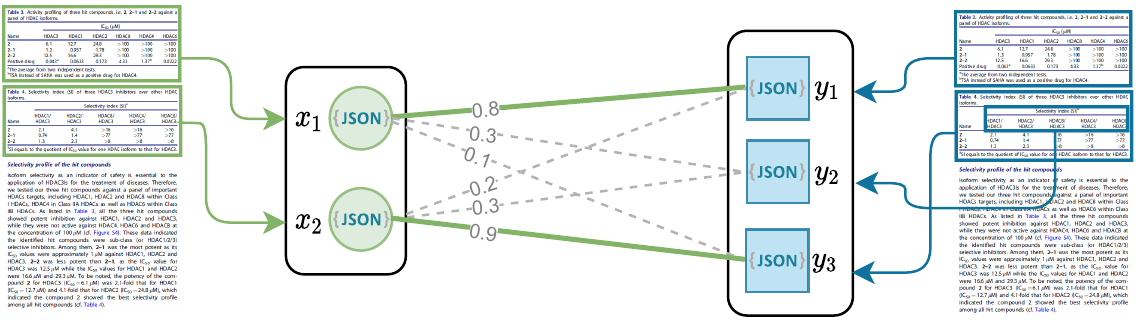}
\caption{A weighted bipartite interpretation graph with two ground-truth and three recognized tables (green circles and blue squares, respectively).
Each vertex corresponds to a set of tuples extracted from a table.
We store each set in a separate JSON file. 
The edge weights are the $F_1$ scores of the matching between the corresponding sets of tuples. 
Green solid lines mark the matching with the maximum sum of weights.
The $y_2$ vertex corresponds to a false-positive result.
}
\label{fig:interpretation-graph}
\end{figure*}

Note that not every table in a document contains information germane to our scenario.
Even if it is the case, not every column has to contain relevant information.
Therefore, we carefully designed the rules employed by our method (\S\ref{sec:interpretation-proposed}) using the development data.
To this end, we employed the fuzzy and RegEx string matching algorithms.
In \Cref{fig:interpr-rules}, we present the exact set of rules that we developed.

\begin{figure}[!tp]
\captionsetup[subfloat]{farskip=1pt,captionskip=1pt}
\centering
\begin{adjustbox}{valign=c}
\verbfilebox[\scriptsize\hspace{0ex}]{../figures/external/rules.txt}\fbox{\theverbbox}
\end{adjustbox}
\par\smallskip 
\caption{
A~JSON file defining the meanings and rules for matching columns to these meanings used in our table interpretation experiment.
}
\label{fig:interpr-rules}
\end{figure}

To evaluate our table extraction method, we first recognized all tables in the evaluation data set using our table recognition algorithms (\S\ref{sec:proposed-method}).
We used the same hyper-parameters as in the ICDAR 2013 experiment (\S\ref{ssec:hyperparams}).
We then employed our interpretation method to extract the relevant tuples from the recognized tables.
To facilitate evaluation, the extracted tuples for each table are stored in a separate JSON file (\S\ref{fig:interpretation_result}).
Moreover, we use the same file name pattern as in the case of the ground-truth files.

The evaluation script takes two sets of JSON files that correspond to the ground-truth and the recognized tables, respectively, as input. 
For every page, we created a bipartite graph with two sets of nodes corresponding to the ground-truth and the recognized tables, respectively (\Cref{fig:interpretation-graph}). 
Subsequently, we performed maximum weighted matching~\citep{Edmonds1965b} to find the correspondence between these two sets of tables.
Finally, we gathered the results from all pages and calculated the exact \emph{precision}, \emph{recall}, and \emph{F$_{1}$} score.
Note that all tuples from the missed reference tables and incorrectly extracted relations were also included in these results.
It is worth noting that these scores reflect the cumulative performance of the complete table extraction process.

{\renewcommand{\arraystretch}{1.0}\setlength{\tabcolsep}{0.0pt}
\newcommand{\ColWidthA}{0.3}
\newcommand{\ColWidthB}{0.06}
\newcommand{\ColWidthC}{0.14}
\newcommand{\ColWidthD}{0.12}
\begin{table*}[!htbp]
\caption{
Results of information extraction from tabular data. 
We include the scores obtained both through the end-to-end table extraction process (\emph{Ours: end-to-end}) and solely from the correctly recognized tables (\emph{Ours: interpretation-only}). 
We report the precision, recall, and F$_{1}$ score.
TP, FP, and FN refer to the number of tuples that were perfectly matched (\emph{true-positive}), missed (\emph{false-negative}), or incorrectly recognized (\emph{false-positive}), respectively.
}
\centering
\begin{tabular}{@{}L{\ColWidthA}*{3}{C{\ColWidthB}}*{2}{C{\ColWidthC}}*{1}{C{\ColWidthD}}@{}}
\toprule
\textbf{Method} & \textbf{TP} & \textbf{FP} & \textbf{FN} & \textbf{Precision} & \textbf{Recall} & \textbf{F$_{1}$}\\
\midrule
Ours: end-to-end              & 69 & 4  & 45 & 0.9452 & 0.6053 & 0.7380\\
Ours: interpretation-only     & 69 & 4  & 5  & 0.9452 & 0.9324 & 0.9388\\
\bottomrule
\end{tabular}
\label{tab:interpretation-results}
\end{table*}}

\subsection{Evaluation Results and Discussion}

\Cref{tab:interpretation-results} presents the results of our method.
We extracted 74 tuples from 10 out of 28 tables.
We achieved a solid complete table extraction F$_{1}$ score of $0.7380$.
Moreover, when we excluded the results from the missed reference tables, our table interpretation method exhibited a high F$_{1}$ score of $0.9388$, proving its utility.
Furthermore, the quantitative analysis revealed that only one false-positive and false-negative error was directly related to the designed interpretation rules.
The remaining errors resulted from table segmentation issues like incorrectly merged cells.

\onlyinsubfile{
\bibliographystyle{../apalike}
\bibliography{../paper}
}

\section{\uppercase{Conclusions}}
\label{sec:conclusions}

In this paper, we presented our flexible and modular table extraction system.
To infer the exact structure of tables in unstructured documents, we developed two heuristics that work with both born-digital and image-based inputs (\S\ref{sec:proposed-method}).
For semantic information extraction, we introduced a configurable graph-based table interpretation method (\S\ref{sec:interpretation-proposed}). 

We conducted extensive experiments on challenging table recognition benchmarks and achieved results that are competitive with state-of-the-art methods (\S\ref{sec:experiments-tablerec}).
In particular, we outperformed other approaches in terms of precision in all evaluation scenarios.

Finally, we evaluated the accuracy of the complete information extraction process and confirmed the utility of our holistic approach (\S\ref{sec:experiments-tableinterpr}).
To foster future research on information extraction from tabular data, we made the evaluation scripts, ground-truth annotations, hyper-parameters, and results of our method publicly available.

We expect that a system \emph{combining} deep learning-based detection and heuristic-based segmentation would further improve the accuracy of complete table recognition.
Therefore, in future work, we integrate a deep learning-based detection module to decrease the precision-recall gap in our results.

Our method has been evaluated on common benchmarks but is not limited to these use cases. Our system is
\begin{enumerate*}
	\item[(1)]{\emph{flexible}, allowing both image-based and digital-born documents,}
	\item[(2)]{\emph{hybrid}, combining heuristics for different layouts,} 
	\item[(3)]{\emph{modular}, covering all processing steps, and allowing to separately adapt the interpretation module to specific scenarios.}
\end{enumerate*}	
Perspectively, we intend to process various documents, such as invoices or balance sheets.


\section*{\uppercase{Acknowledgments}}
This work was supported by the Fraunhofer Internal Programs under Grant No. 836 885.


\bibliographystyle{apalike}

{\small
\bibliography{main}
}

\appendix

\section{\uppercase{Appendix}}
\label{sec:app-interpr}

In this section, we present an example of a ground-truth file from our data set (\Cref{fig:interpr-gt}) that we used to evaluate table interpretation (cf. \S\ref{sec:experiments-tableinterpr}).

\begin{figure}[!htbp]
\captionsetup[subfloat]{farskip=1pt,captionskip=1pt}
\centering
\begin{adjustbox}{valign=c}
\verbfilebox[\scriptsize\hspace{0ex}]{../figures/external/gt.txt}\fbox{\theverbbox}
\end{adjustbox}
\par\smallskip 
\caption{
An example of a ground-truth file from our collection used in our table interpretation experiment (\emph{11\_page07\_table0.json}).
}
\label{fig:interpr-gt}
\end{figure}

\end{document}